# Basic Reasoning with Tensor Product Representations

**Paul Smolensky**[*]
Department of Cognitive Science
Johns Hopkins University
Baltimore, MD 21218, USA
`smolensky@jhu.edu`

**Moontae Lee**
Department of Computer Science
Cornell University
Ithaca, NY 14850, USA
`moontae@cs.cornell.edu`

**Xiaodong He, Wen-tau Yih, Jianfeng Gao & Li Deng**
Microsoft Research
Redmond, WA 98052, USA
`{xiaohe, scottyih, jfgao, deng}@microsoft.com`

In this paper we present the initial development of a general theory for mapping inference in predicate logic to computation over Tensor Product Representations (TPRs; Smolensky (1990), Smolensky & Legendre (2006)). After an initial brief synopsis of TPRs (Section 0), we begin with particular examples of inference with TPRs in the 'bAbI' question-answering task of Weston et al. (2015) (Section 1). We then present a simplification of the general analysis that suffices for the bAbI task (Section 2). Finally, we lay out the general treatment of inference over TPRs (Section 3). We also show the simplification in Section 2 derives the inference methods described in Lee et al. (2016); this shows how the simple methods of Lee et al. (2016) can be formally extended to more general reasoning tasks.

## 0    BRIEF SYNOPSIS OF TPR

For present purposes, a tensor $\mathbf{T}^{(n)}$ of order $n$ over $\mathbb{R}^d$ can be taken to be an $n$-dimensional array of real numbers, each written $\mathbf{T}_{\gamma_1 \cdots \gamma_n}$, $\gamma_k \in \{1, 2, \ldots, d\} \equiv 1{:}d$ for all $k \in 1{:}n$. The two types of tensor operations we use are given in (1): the outer or tensor product (1a) is order-increasing, while contraction (1b) is order-decreasing. Combining the two gives the inner product (1c). If we interpret an order-2 tensor $\mathbf{M}^{(2)}$ as a matrix $\mathbf{M}$, and order-1 tensors $\mathbf{U}^{(1)}$, $\mathbf{V}^{(1)}$ as vectors/column-matrices $\mathbf{u}$, $\mathbf{v}$, then the outer product $\mathbf{u}\mathbf{v}^\mathsf{T}$ of matrix algebra corresponds to the tensor product $\mathbf{u} \otimes \mathbf{v}$ (1d) while the dot product $\mathbf{u} \cdot \mathbf{v} = \mathbf{u}^\mathsf{T}\mathbf{v}$ and the matrix-vector product $\mathbf{M}\mathbf{u}$ correspond to tensor inner products (1e).

---

[*] This work was conducted while the first author was a Visiting Researcher, and the second author held a summer internship, at Microsoft Research, Redmond, WA.

(1) Tensor operations

    a. outer/tensor product    $\mathbf{U}^{(n)} \otimes \mathbf{V}^{(m)} = \mathbf{T}^{(n+m)}$    $\mathbf{T}_{\gamma_1 \cdots \gamma_n \gamma'_1 \cdots \gamma'_m} \equiv \mathbf{U}_{\gamma_1 \cdots \gamma_n} \mathbf{V}_{\gamma'_1 \cdots \gamma'_m}$

    b. contraction    $\mathfrak{C}_{jk}\mathbf{U}^{(n)} = \mathbf{T}^{(n-2)}$    $\mathbf{T}_{\gamma_1 \cdots \gamma_{j-1}\gamma_{j+1} \cdots \gamma_{k-1}\gamma_{k+1} \cdots \gamma_n} \equiv \Sigma_\beta \mathbf{T}_{\gamma_1 \cdots \gamma_{j-1}\beta\gamma_{j+1} \cdots \gamma_{k-1}\beta\gamma_{k+1} \cdots \gamma_n}$

    c. inner product [$j < n < k$]   $\mathbf{U}^{(n)} \bullet_{jk} \mathbf{V}^{(m)} = \mathbf{T}^{(n+m-2)}$   $\mathbf{T} \equiv \mathfrak{C}_{jk}\mathbf{U} \otimes \mathbf{V}$

$$\therefore \mathbf{T}_{\gamma_1 \cdots \gamma_{j-1}\gamma_{j+1} \cdots \gamma_{k-1}\gamma_{k+1} \cdots \gamma_{n+m}} = \Sigma_\beta \mathbf{U}_{\gamma_1 \cdots \gamma_{j-1}\beta\gamma_{j+1} \cdots \gamma_n} \mathbf{V}_{\gamma_{n+1} \cdots \gamma_{k-1}\beta\gamma_{k+1} \cdots \gamma_{n+m}}$$

    d. $[\mathbf{U}^{(1)} \otimes \mathbf{V}^{(1)}]_{\gamma\gamma'} = \mathbf{U}_\gamma \mathbf{V}_{\gamma'} \cong \mathbf{u}_\gamma \mathbf{v}_{\gamma'} = [\mathbf{uv}^\mathsf{T}]_{\gamma\gamma'}$

    e. $\mathbf{U}^{(1)} \cdot \mathbf{V}^{(1)} \equiv \mathbf{U}^{(1)} \bullet_{12} \mathbf{V}^{(1)} = \mathbf{U}_\beta \mathbf{V}_\beta \cong \mathbf{u}_\beta \mathbf{v}_\beta = \mathbf{u} \cdot \mathbf{v}$;   $[\mathbf{M}^{(2)} \bullet_{23} \mathbf{U}^{(1)}]_\gamma = \mathbf{M}_{\gamma\beta}\mathbf{U}_\beta \cong \mathbf{M}_{\gamma\beta}\mathbf{u}_\beta = [\mathbf{Mu}]_\gamma$

Following the customary practice, throughout the paper, except where explicitly stated otherwise, we assume an implicit summation over repeated indices in a single factor — the Einstein Summation Convention. Thus the explicit summation over β in (1b–c) would be omitted and left implicit, as in (1e).

A particular TPR maps a space $\mathcal{S}$ of symbolic structures to a vector space $\mathbb{R}^N$. The type of a structure $s \in \mathcal{S}$ is determined by a set $\mathcal{R} = \{r_k\}$ of *structural roles* that determines a *filler/role decomposition b* of $\mathcal{S}$: each token structure $s$ is uniquely characterized as a set of *filler/role bindings* $b(s) = \{\mathtt{f}_k/r_k\}$, in which each role $r_k$ is bound to a particular filler $\mathtt{f}_k \in \mathcal{F}$. As an illustration of one type of filler/role decomposition, *positional roles*, let $\mathcal{S}$ be the set of strings over the alphabet of symbols $\mathcal{A} = \{\mathtt{a, b, c}\}$ and let $r_k$ be the role of the $k^{\text{th}}$ symbol (from the left). The $\mathcal{F} = \mathcal{A}$ and $\mathcal{R} = \{r_1, r_2, \ldots\}$. For the particular string $\mathtt{acb}$, we have $b_{\text{pos}}(\mathtt{acb}) = \{\mathtt{a}/r_1, \mathtt{b}/r_3, \mathtt{c}/r_2\}$; note that the bindings constitute a *set*. To illustrate the other type of filler/role decomposition, *contextual roles*, for the same type of structure $\mathcal{S}$, strings, let $r_{x\_y}$ denote the role 'preceded by symbol $x$ and followed by symbol $y$'. Then in this new decomposition $b_{\text{con}}$, $\mathtt{acb}$ has only one binding: $b_{\text{con}}(\mathtt{acb}) = \{\mathtt{c}/r_{\mathtt{a\_b}}\}$. In this decomposition, a string is characterized by its trigrams.

Given a filler/role decomposition $b$ for $\mathcal{S}$, a TPR is defined by encoding each filler $\mathtt{f}_k \in \mathcal{F}$ by a *filler tensor* $\mathbf{f}_k \in V_\mathcal{F}$, and each role $r_k \in \mathcal{R}$ by a *role tensor* $\mathbf{r}_k \in V_\mathcal{R}$. Role tensors are a principal innovation of TPR.

Then the TPR of a structure $s$ with bindings $b(s) = \{\mathtt{f}_k/r_k\}$ is the tensor $\mathbf{s} \equiv \Sigma_k \mathbf{f}_k \otimes \mathbf{r}_k \in V_\mathcal{S} \equiv V_\mathcal{F} \otimes V_\mathcal{R} \equiv \{\mathbf{f} \otimes \mathbf{r} \mid \mathbf{f} \in V_\mathcal{F}, \mathbf{r} \in V_\mathcal{R}\}$. Thus in TPR, binding is done via the tensor product. For the positional-role decomposition $b_{\text{pos}}$, the TPR of $s = \mathtt{acb}$ is $\mathbf{s}_{\text{pos}} = \mathbf{a} \otimes \mathbf{r}_1 + \mathbf{b} \otimes \mathbf{r}_3 + \mathbf{c} \otimes \mathbf{r}_2$. We use this type of positional-role TPR below in (18).

Now consider the contextual-role decomposition $b_{\text{con}}$ of $\mathcal{S}$. Because $b_{\text{con}}(\mathtt{acb}) = \{\mathtt{c}/r_{\mathtt{a\_b}}\}$, the TPR of $s = \mathtt{acb}$ is $\mathbf{s}_{\text{con}} = \mathbf{c} \otimes \mathbf{r}_{\mathtt{a\_b}}$. We need a tensor to encode each role $r_{x\_y}$. Such a role is itself a structure, which can be given a filler/role decomposition such that $r_{x\_y}$ is the binding $x/r_{\_\_y}$, giving rise to the encoding tensor $\mathbf{r}_{x\_y} \equiv \mathbf{x} \otimes \mathbf{r}_{\_\_y}$. For the role tensor $\mathbf{r}_{\_\_y}$ we can choose the filler vector $\mathbf{y}$, so $\mathbf{r}_{x\_y} = \mathbf{x} \otimes \mathbf{y}$. Then the TPR for $s = \mathtt{acb}$ is $\mathbf{s}_{\text{con}} = \mathbf{c} \otimes [\mathbf{a} \otimes \mathbf{b}]$. An isomorphic encoding is the more mnemonic $\mathbf{s}_{\text{con}} = \mathbf{a} \otimes \mathbf{c} \otimes \mathbf{b}$. Thus here the role tensors in $V_\mathcal{R}$ are of order 2, and the vector encoding a string is a tensor of order 3.

Our primary interest is in vectorial encodings of propositions such as $P(\mathtt{a, b, c})$. We will adopt a contextual TPR such that the encoding of this proposition is $\mathbf{P} \otimes \mathbf{a} \otimes \mathbf{b} \otimes \mathbf{c}$. The corresponding TPR encoding of a set of propositions $\mathcal{B} = \{\mathtt{P}_k(\mathtt{a}_k, \mathtt{b}_k, \mathtt{c}_k)\}$ is $\mathbf{B} = \Sigma_k \mathbf{P}_k \otimes \mathbf{a}_k \otimes \mathbf{b}_k \otimes \mathbf{c}_k$. The space of such order-4 tensors $\mathbf{B}$ is a vector space of dimension $4d$, assuming each symbol is encoded by an order-1 tensor over $\mathbb{R}^d$, i.e., for each symbol $\mathtt{a}$ the components of its tensor encoding are $[\mathbf{a}]_\gamma$, $\gamma \in 1{:}d$.

We will assume that the order-1 tensors $\mathbf{f}_k$ chosen to encode the symbols $\mathtt{f}_k$ form an orthonormal set: $\mathbf{f}_j \cdot \mathbf{f}_k = \delta_{jk} \equiv [1 \text{ IF } j = k \text{ ELSE } 0]$. This assures that TPRs can be *unbound* with perfect accuracy, via the inner product, which 'undoes' the outer product binding. As an example, consider a set of propositions $\mathcal{B} = \{\mathtt{P}_k(\mathtt{a}_k, \mathtt{b}_k, \mathtt{c}_k)\}$, only one of which has the form $\mathtt{P}_2(\mathtt{a}_2, \mathtt{b}_2, x)$. We can find the unique value of $x$ (namely $\mathtt{c}_2$) such that $\mathtt{P}_2(\mathtt{a}_2, \mathtt{b}_2, x) \in \mathcal{B}$ from $\mathcal{B}$'s TPR encoding $\mathbf{B} = \Sigma_k \mathbf{P}_k \otimes \mathbf{a}_k \otimes \mathbf{b}_k \otimes \mathbf{c}_k$, by computing $\mathbf{x} = \mathbf{B} \bullet_{15,26,37} (\mathbf{P}_2 \otimes \mathbf{a}_2 \otimes \mathbf{b}_2)$, i.e.,

$$\mathbf{x}_\gamma = \mathbf{B}_{\pi\alpha\beta\gamma}[\mathbf{P}_2]_\pi[\mathbf{a}_2]_\alpha[\mathbf{b}_2]_\beta = \left(\Sigma_k [\mathbf{P}_k]_\pi[\mathbf{a}_k]_\alpha[\mathbf{b}_k]_\beta[\mathbf{c}_k]_\gamma\right)[\mathbf{P}_2]_\pi[\mathbf{a}_2]_\alpha[\mathbf{b}_2]_\beta = \Sigma_k [\mathbf{P}_k \cdot \mathbf{P}_2][\mathbf{a}_k \cdot \mathbf{a}_2][\mathbf{b}_k \cdot \mathbf{b}_2][\mathbf{c}_k]_\gamma = [\mathbf{c}_2]_\gamma$$

because for every value of $k$ except $k = 2$, either $\mathbf{P}_k \neq \mathbf{P}_2$ or $\mathbf{a}_k \neq \mathbf{a}_2$ or $\mathbf{b}_k \neq \mathbf{b}_2$, so $[\mathbf{P}_k \cdot \mathbf{P}_2][\mathbf{a}_k \cdot \mathbf{a}_2][\mathbf{b}_k \cdot \mathbf{b}_2] = 0$.

*Comment.* While we assume for convenience throughout the paper that the tensors encoding symbols are orthonormal, there is no assumption that they are 1-hot vectors; we presume they are distributed vectors in which many elements are non-zero. Further, all results here would continue to hold if the tensors encoding symbols were merely linearly independent; unbinding would then be done with *unbinding filler tensors* $\mathbf{f}_k^+$ replacing the filler tensors $\mathbf{f}_k$ themselves. The unbinding example of the previous paragraph would become $\mathbf{x} = \mathbf{B} \bullet_{15,26,37} (\mathbf{P}_2^+ \otimes \mathbf{a}_2^+ \otimes \mathbf{b}_2^+)$, where the unbinding tensors $\mathbf{f}_k^+$ are defined so that $\mathbf{f}_j \cdot \mathbf{f}_k^+ = \delta_{jk}$. Such vectors must exist if the filler tensors $\{\mathbf{f}_k\}$ are linearly independent (essentially, $\mathbf{f}_k^+$ is the $k^{\text{th}}$ row of the inverse of the matrix $\mathbb{F}$ which has $\mathbf{f}_k$ as its $k^{\text{th}}$ column; $\mathbb{F}$ is invertible if the $\{\mathbf{f}_k\}$ are linearly independent).

# 1    BABI EXAMPLE

Consider the example in (2). "@($a$, $b$, $t$)" denotes "$a$ is at $b$ at time $t$" (or "$a$ is co-located with $b$ at time $t$"). (In Lee et al. (2016), the gloss is "$a$ belongs to $b$" or "$a$ is contained in $b$".) ⓞ is the information-question operator; assuming that the denotation of a question is the set of answers, ⓞ$x.P(x)$ denotes "the $x$'s for which it the case that $P(x)$" = $\{x \mid P(x)\}$. The English question "where was the apple before the kitchen?" is assigned a Logical Form (LF) that can be glossed as "the location $x$ for which it is the case that a [the apple] was at $x$ at some time $t$ and the apple was at k [the kitchen] at the time $t'$ immediately following $t$".

(2)    An example of a type-3 question from the bAbI task:

| | | |
|---|---|---:|
| a. | John picked up an apple | @(a, j, $t_1$) |
| b. | John went to the office | @(j, f, $t_2$) |
| c. | John went to the kitchen | @(j, k, $t_3$) |
| d. | John dropped the apple | ¬@(a, j, $t_4$) |
| e. | Where was the apple before the kitchen? | ⓞ$x.\exists t, t'$. @(a, $x$, $t$) |
| | → the office | & @(a, k, $t'$) |
| | | & ≺($t$, $t'$) |

Here we will assume given a [surface string → LF] semantic parser that generates the right column of (2) given the left column. We strive to separate issues of commonsense inference per se from issues of NLP narrowly construed, such as identifying: the semantic predicates corresponding to English words, the referents of referring expressions, the antecedents of anaphoric expressions, and the content of elided material. We thus assume given NLP procedures for performing such computations and focus exclusively on the problem of commonsense reasoning with distributed, vectorial representations.

All symbols in the predicate logic analysis are encoded in TPR as vectors, or order-1 tensors, in $\mathbb{R}^d$. Thus **@** is a vector encoding the symbol '@', and $@_\pi \in \mathbb{R}$, $\pi = 1{:}d$, are its $d$ real components. In general, the symbol-encoding vectors such as **@** are distributed, in the sense that many components are non-zero (they are not in general 1-hot vectors). For convenience, here we assume these vectors are an orthonormal set (what is actually required is only that they be linearly independent).

At each time $t_i$, the reasoning process is in a state $\mathcal{B}(t_i)$ which we take to be the set of propositions constituting the knowledge base of facts concerning the problem situation; this grows monotonically with $t_i$, as more information arrives: the LF form of the $i^{\text{th}}$ sentence, L($t_i$). "≺($t$, $t'$)" denotes the proposition "the time $t$ immediately precedes the time $t'$". As illustrated in (2), we will often notate times as "$t_i$", $i \in \mathbb{N}$, where $\forall i. \prec(t_i, t_{i+1})$. The times $\{t_i\}$ have TPRs $\{\mathbf{t}_i\} \subset \mathbb{R}^d$ that are linearly independent, so there is a linear operator $\mathbb{T}$ on $\mathbb{R}^d$ satisfying (3).

(3) Time-increment operator $\mathbb{T}$

$$\mathbb{T}\mathbf{t}_i = \mathbf{t}_{i+1}$$

The TPR of the time-$t_i$ knowledge base $\mathcal{B}(t_i)$ is the tensor $\mathbf{B}(t_i)$. This fourth-order tensor is the sum of the TPRs of propositions of the form @($x$, $y$, $t$) or ≺($t$, $t'$, ø); ø is a dummy symbol used for convenience to make ≺, like @, a predicate that takes 3 arguments. The propositions are given a contextual TPR:

(4) TPR of propositions

    a. @($x$, $y$, $t$)      **@** ⊗ **x** ⊗ **y** ⊗ **t**

    b. ≺($t$, $t'$, ø)      **≺** ⊗ **t** ⊗ **t'** ⊗ **ø**

The four indices in $\mathbf{B}_{\pi\alpha\beta\tau}$ can be thought of as the proposition-, first-argument-, second-argument-, and third-argument-indices. For a proposition with predicate @ such as @(j, k, $t$) (2c) or @(a, j, $t'$) (2a), $\alpha$ is the index of an actor- or object-vector, $\beta$ is the index of a location- or actor-vector, and $\tau$ is the index of a time-vector: the TPR of such proposition is $\mathbf{b}_{\pi\alpha\beta\tau} = \mathbf{@}_\pi \mathbf{j}_\alpha \mathbf{k}_\beta \mathbf{t}_\tau$ or $\mathbf{@}_\pi \mathbf{a}_\alpha \mathbf{j}_\beta \mathbf{t'}_\tau$.

The reasoning in example (2) requires two rules of inference:

(5) Transitivity Axiom for @

$$\forall x,y,z,t.\; @(x, y, t)\; \&\; @(y, z, t) \Rightarrow @(x, z, t)$$

(6) Persistence Axiom for @

$$\forall x,y,t,t'.\; @(x, y, t)\; \&\; \prec(t,t') \Rightarrow @(x, y, t')$$

In the vectorial reasoning system we develop, the persistence axiom can be applied at every time $t$, deriving positions at the immediately following time $t'$. Expressions like (2d) "John dropped the apple" are interpreted as ¬@(a, j, $t$) "not: the apple is at John [at time $t$]", and the tensor **b** encoding this proposition will be the negation of the vector encoding "the apple is at John"; **b** will simply cancel the vector for "the apple is at John" that was generated by the persistence axiom. Then at subsequent times there is no longer an encoded proposition @(a, j, $t$) that the persistence axiom can propagate forward.

The reasoning needed for (2) can be expressed as

    $t_1$: the <u>a</u>pple is at John

    $t_2$: ⇒$_\text{persistence}$ the <u>a</u>pple is at John, John is at the <u>o</u>ffice    ⇒$_\text{transitivity}$ the <u>a</u>pple is at the <u>o</u>ffice

    $t_3$: ⇒$_\text{persistence}$ the <u>a</u>pple is at John, John is at the <u>k</u>itchen  ⇒$_\text{transitivity}$ the <u>a</u>pple is at the <u>k</u>itchen

    $t_4$: ⇒$_\text{persistence}$ the <u>a</u>pple is at the <u>k</u>itchen, ~~the apple is at John~~

The TP encodings of the inference rules (5)–(6) are given in (7)–(8); these encodings are derived by a general procedure that will be illustrated in Section 3.

(7) TP encoding of the Transitivity Axiom for @

   a. $\forall x,y,z,t.\quad @(x, z, t) \Leftarrow \quad @(x, y, t)\ \&\ @(y', z, t)\ \&\ [y = y']$
   b. $[\mathfrak{V}(\mathbf{B}(t))]_{\pi''\alpha\beta'\tau''} = @_{\pi''}\mathbf{t}_{\tau''}[\mathbf{B}(t)]_{\pi\alpha\beta\tau}\,@_{\pi}\mathbf{t}_{\tau}\,[\mathbf{B}(t)]_{\pi'\alpha'\beta'\tau'}\,@_{\pi'}\mathbf{t}_{\tau'}\,\delta_{\beta\alpha'} + [\mathbf{B}(t)]_{\pi''\alpha\beta'\tau''}$
   c. $\mathfrak{V}(\mathbf{B}(t)) = \mathbb{V}[\mathbf{B}(t), \mathbf{B}(t); t]$
   d. $\mathbb{V}$ = a multilinear tensor operation encoding inference from the Transitivity Axiom:
      i. $\forall x,y,z,t; \forall p \in T.\ p(x, z, t) \Leftarrow\ p(x, y, t)\ \&\ p(y, z, t)$
      ii. $\mathbb{V}[\mathbf{X}, \mathbf{Y}; t]_{\pi''\alpha\beta'\tau''} = \sum_{p \in T}\mathbf{p}_{\pi''}\bigl(\mathbf{p}_{\pi}[\mathbf{X}]_{\pi\alpha\beta\tau}\mathbf{t}_{\tau}\bigr)\bigl(\mathbf{p}_{\pi'}[\mathbf{Y}]_{\pi'\beta\beta'\tau'}\mathbf{t}_{\tau'}\bigr)\mathbf{t}_{\tau''}$
      iii. (Modified) Penrose diagram (for one $p \in T$):

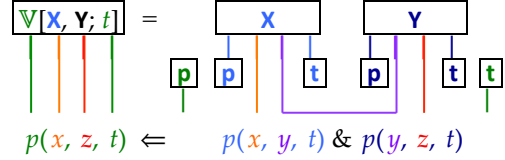

   $p(x, z, t) \Leftarrow\quad p(x, y, t)\ \&\ p(y, z, t)$

(8) TP encoding of the Persistence Axiom for @ [note that $\forall i.\ \mathbb{T}\mathbf{t}_{i-1} = \mathbf{t}_i$, so $\prec(t,t')$ iff $\mathbf{t}' = \mathbb{T}\mathbf{t}$ iff $\mathbf{t} = \mathbb{T}^{-1}\mathbf{t}'$]

   a. $\forall x,y,t,t'.\qquad @(x, y, t') \Leftarrow @(x, y, t)\ \&\ \prec(t,t')$
   b. $\mathbf{B}(t_i) = \bigl(\mathbb{1} + \sum_{x,y}[@ \otimes \mathbf{x} \otimes \mathbf{y} \otimes \mathbf{t}_i][@ \otimes \mathbf{x} \otimes \mathbf{y} \otimes \mathbb{T}^{-1}\mathbf{t}_i]^\top\bigr)\mathbf{B}(t_{i-1})$
   c. $\mathbf{B}(t_i) = \bigl(\mathbb{1} + \mathbb{P}(t_i)\bigr)\mathbf{B}(t_{i-1})$
   d. $\mathbb{P}(t)$ = matrix operating on $\mathbf{B}$-tensors that encodes inference from the Persistence Axiom for @
      i. $[\mathbb{P}(t_i)\,\mathbf{B}(t_{i-1})]_{\pi\xi\eta\tau} = \mathbb{P}(t)_{\pi\xi\eta\tau,\,\pi'\xi'\eta'\tau'}\,[\mathbf{B}(t_{i-1})]_{\pi'\xi'\eta'\tau'}$
      ii. $\mathbb{P}(t)_{\pi\xi\eta\tau,\,\pi'\xi'\eta'\tau'}\quad =\quad @_{\pi}\,\delta_{\xi\xi'}\,\delta_{\eta\eta'}\,\mathbf{t}_{\tau}\ @_{\pi'}\quad [\mathbb{T}^{-1}]_{\tau'\tau''}\,\mathbf{t}_{\tau''}$
      iii. Penrose diagram:

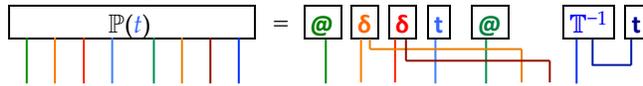

Here and throughout the paper, except where explicitly stated otherwise, we deploy the Einstein Summation Convention, according to which repeated indices are implicitly summed over all their values; e.g., in (7b) there is an implicit sum over $\pi$, $\beta$, $\tau$, $\pi'$, $\alpha'$, and $\tau'$. In (7b) and (8d.ii), $\delta$ is the Kronecker $\delta$: $\delta_{ij} \equiv$ [1 IF $i=j$ ELSE 0].

In the general case (7d.ii), the encoding of inference using the Transitivity Axiom involves a sum over the set $T$ of all transitive predicates; we will only be using the single transitive predicate @ here, and the other expressions in (7)–(8) deal only with that predicate, which is persistent as well as transitive.

In (modified) Penrose Tensor Diagrams such as (7d.iii) and (8d.iii), each box denotes a tensor and the $n^{\text{th}}$ line from the left that emanates from the box for tensor $\mathbf{A}$ denotes the $n^{\text{th}}$ index of $\mathbf{A}$; there are $m$ such lines if $\mathbf{A}$ is an $m^{\text{th}}$-order tensor. When two lines are joined, the values of those two indices are set equal and there is sum over all values for the index, as explicitly shown for $\beta$ in (7d.ii), the algebraic expression denoted by the Penrose Diagram (7d.iii). Penrose Diagrams enable complex tensor equations to be written precisely without any indices, thereby making the structure of the equations more transparent.

The correspondence between a predicate logic expression and a Penrose Tensor Diagram will be made explicit in Section 3, but the juxtaposition in (7d.iii) of the Penrose diagram and the corresponding predicate logic expression beneath it already suggests the nature of the correspondence visually, with colors of predicate logic symbols matching the colors of their corresponding tensors in the diagram as well as the colors of their corresponding indices in the algebraic expressions (7d.ii), (8d.i).

In (8b)–(8c), the identity matrix over tensors, $\mathbb{1}$, ensures that all the propositions encoded in **B** at time $t_{i-1}$ (propositions about the problem situation at all times $t \le t_{i-1}$) are carried over and also encoded in **B** at time $t_i$.

An example of the use of the Transitive Inference procedure of (7) is given in (9). (Recall that the TPRs of all symbols form an orthonormal set.)

(9) Example of Transitivity Inference using (7)

    a. Let $\mathcal{B}(t_2) = \{@(a, j, t_1), @(a, j, t_2), @(j, k, t_2)\}$; result of transitive inference: add $@(a, k, t_2)$

    b. Then $\mathbf{B}(t_2) = @ \otimes \mathbf{a} \otimes \mathbf{j} \otimes \mathbf{t}_1 + @ \otimes \mathbf{a} \otimes \mathbf{j} \otimes \mathbf{t}_2 + @ \otimes \mathbf{j} \otimes \mathbf{k} \otimes \mathbf{t}_2$

    c. $\mathbf{B}(t_2) \otimes \mathbf{B}(t_2) = @ \otimes \mathbf{a} \otimes \mathbf{j} \otimes \mathbf{t}_1 \otimes @ \otimes \mathbf{j} \otimes \mathbf{k} \otimes \mathbf{t}_2 + @ \otimes \mathbf{a} \otimes \mathbf{j} \otimes \mathbf{t}_2 \otimes @ \otimes \mathbf{j} \otimes \mathbf{k} \otimes \mathbf{t}_2 + \cdots$

    d. $[\mathfrak{V}(\mathbf{B}(t_2))]_{\pi''\alpha\beta'\tau''} = @_{\pi''}\mathbf{t}_{2\tau''}[\mathbf{B}(t_2)]_{\pi\alpha\beta\tau} @_{\pi} \mathbf{t}_{2\tau} [\mathbf{B}(t_2)]_{\pi'\alpha'\beta'\tau'} @_{\pi'}\mathbf{t}_{2\tau'}\delta_{\beta\alpha'} + [\mathbf{B}(t_2)]_{\pi''\alpha\beta'\tau''}$

$\qquad\qquad = @_{\pi''}\mathbf{t}_{2\tau''}[@ \otimes \mathbf{a} \otimes \mathbf{j} \otimes \mathbf{t}_1 \otimes @ \otimes \mathbf{j} \otimes \mathbf{k} \otimes \mathbf{t}_2]_{\pi\alpha\beta\tau\pi'\alpha'\beta'\tau'} @_{\pi}\mathbf{t}_{2\tau}@_{\pi'}\mathbf{t}_{2\tau'}\delta_{\beta\alpha'}$

$\qquad\qquad + @_{\pi''}\mathbf{t}_{2\tau''}[@ \otimes \mathbf{a} \otimes \mathbf{j} \otimes \mathbf{t}_2 \otimes @ \otimes \mathbf{j} \otimes \mathbf{k} \otimes \mathbf{t}_2]_{\pi\alpha\beta\tau\pi'\alpha'\beta'\tau'} @_{\pi}\mathbf{t}_{2\tau}@_{\pi'}\mathbf{t}_{2\tau'}\delta_{\beta\alpha'} + \cdots + [\mathbf{B}(t_2)]_{\pi''\alpha\beta'\tau''}$

$\qquad\qquad = @_{\pi''}\mathbf{t}_{2\tau''}[@_{\pi}@_{\pi}]\mathbf{a}_{\alpha}[\mathbf{j}_{\beta} \mathbf{j}_{\alpha'}\delta_{\beta\alpha'}][\mathbf{t}_{1\tau}\mathbf{t}_{2\tau}][@_{\pi'}@_{\pi'}]\mathbf{k}_{\beta'}[\mathbf{t}_{2\tau}\mathbf{t}_{2\tau'}]$

$\qquad\qquad + @_{\pi''}\mathbf{t}_{2\tau''}[@_{\pi}@_{\pi}]\mathbf{a}_{\alpha}[\mathbf{j}_{\beta} \mathbf{j}_{\alpha'}\delta_{\beta\alpha'}][\mathbf{t}_{2\tau}\mathbf{t}_{2\tau}][@_{\pi'}@_{\pi'}]\mathbf{k}_{\beta'}[\mathbf{t}_{2\tau}\mathbf{t}_{2\tau'}] + \cdots + [\mathbf{B}(t_2)]_{\pi''\alpha\beta'\tau''}$

$\qquad\qquad = @_{\pi''}\mathbf{t}_{2\tau''}\mathbf{a}_{\alpha} \mathbf{k}_{\beta'} + [\mathbf{B}(t_2)]_{\pi''\alpha\beta'\tau''} = [@ \otimes \mathbf{a} \otimes \mathbf{k} \otimes \mathbf{t}_2]_{\pi''\alpha\beta'\tau''} + [\mathbf{B}(t_2)]_{\pi''\alpha\beta'\tau''}$

    e. i.e., $\mathfrak{V}(\mathbf{B}(t_2))$ is the TPR of: $@(a, k, t_2) \cup \mathcal{B}(t_2)$

In the evaluation of $\mathfrak{V}(\mathbf{B}(t_2))$ in (9d), because the TPRs of all symbols are orthogonal, all terms in $\mathbf{B}(t_2) \otimes \mathbf{B}(t_2)$ (with components $[\mathbf{B}(t_2)]_{\pi\alpha\beta\tau}[\mathbf{B}(t_2)]_{\pi'\alpha'\beta'\tau'}$) are annihilated except the single term which is the TPR of the proposition pair $\langle @(a, j, t_2), @(j, k, t_2)\rangle$, because the inner product with $@_{\pi}\mathbf{t}_{2\tau}@_{\pi'}\mathbf{t}_{2\tau'}\delta_{\beta\alpha'}$ gives 0 for any pair $\langle p(x, y, t), p'(z, w, t')\rangle$ unless $p = @ = p'$, $y = z$, and $t = t_2 = t'$; e.g., the factor in red brackets $[\mathbf{t}_{1\tau}\mathbf{t}_{2\tau}]$ is 0 because $\mathbf{t}_1$ and $\mathbf{t}_2$ are orthogonal. (The factors $[\mathbf{v}_{\mu}\mathbf{v}_{\mu}]$ all equal 1 because the TPRs of all symbols are normalized to length 1.)

At each consecutive time $t_i$ we also have the following update rule (10) for the (immediate) Temporal Precedence relation $\prec$ (recall the definition of the time-increment operator $\mathbb{T}$ (3)).

(10) Update rule for the symbolic Temporal Precedence relation $\mathcal{T} = \prec$ and its TPR **T**

    a. $\mathcal{T}(t_i) = \prec(t_{i-1}, t_i) \cup \mathcal{T}(t_{i-1})$

    b. $\mathbf{T}(t_i) = \prec \otimes \mathbf{t}_{i-1} \otimes \mathbf{t}_i + \mathbf{T}(t_{i-1}) = \prec \otimes \mathbf{t}_{i-1} \otimes \mathbb{T}\mathbf{t}_{i-1} + \mathbf{T}(t_{i-1})$

The procedure for building the knowledge base **B** incrementally, as the sentence $S_i$ pertaining to each time $t_i$ is processed, is given in (11).

(11) TP Reasoning Algorithm

    a. Goal: knowledge base $\mathcal{B}$      $\mathbf{B} = @ \otimes \mathbf{a} \otimes \mathbf{J} \otimes \mathbf{t}_1 + \prec \otimes \mathbf{t}_1 \otimes \mathbf{t}_2 \otimes \emptyset + \cdots$

         To construct $\mathcal{B}(t_i) / \mathbf{B}(t_i)$, loop over sentences $i$ in story:

    b. $\mathcal{B}(t_{i-1})$ given      $\mathbf{B}(t_{i-1})$ already computed

    c. inferences from Persistence Axiom      $\mathbf{B}(t_i) \leftarrow \mathbf{B}(t_{i-1}) + \mathbb{P}(t_i) \mathbf{B}(t_{i-1})$

$\qquad\qquad \forall k, \forall p \in \mathcal{P}. \; p(a_1, \ldots, a_m; t_{i-1})$      $\mathbb{P}(t_k) =$ Persistence matrix

$\qquad\qquad \Rightarrow p(a_1, \ldots, a_m; t_k)$      (over tensors)

    d. update $\prec$      $\mathbf{B}(t_i) \leftarrow \mathbf{B}(t_i) + \mathbf{T}_i$

$\qquad\qquad$ add $\prec(t_{i-1}, t_i)$      $\mathbf{T}_i = \prec \otimes \mathbf{t}_{i-1} \otimes \mathbb{T}\mathbf{t}_{i-1}$

$\qquad\qquad\qquad\qquad\qquad\qquad\qquad\qquad$ $\mathbb{T} =$ time-update matrix; $\mathbf{t}_i = \mathbb{T}\mathbf{t}_{i-1}$

| | | | | | |
|---|---|---|---|---|---|
| e. | add L($t_i$) = LF of $i^{th}$ sentence | | | $\mathbf{B}(t_i) \leftarrow \mathbf{B}(t_i) + \mathbf{L}(t_i)$ | |
| | e.g., John picked up an apple | | | e.g. $@ \otimes \mathbf{A} \otimes \mathbf{J} \otimes \mathbf{t}_1 = \mathbf{L}(t_1)$ | |
| f. | repeat until no change: | | | | |
| g. | inferences from Transitivity Axiom: | | | $\mathbf{B}(t_i) \leftarrow \mathbf{B}(t_i) + \mathbb{V}[\mathbf{B}(t_i), \mathbf{B}(t_i); t_i]$ | |
| | $\forall x,y,z,t,p \in \mathcal{T}.\ p(x, y, t)\ \&\ p(y, z, t)$ | | | $\mathbb{V}$ = multilinear tensor operation of | |
| | $\Rightarrow p(x, z, t)$ | | | Transitive Inference | |

This algorithm, processing the example (2), will produce (12).

(12) Algorithm (11) processing example (2)

| | Sentence $i$ | LF: L($i$) | Inferences | $\mathcal{T}$ update | Explanation |
|---|---|---|---|---|---|
| a. | John picked up an apple | @(a, j, $t_1$) | | | |
| b. | John went to the office | @(j, f, $t_2$) | @(a, j, $t_2$) | $\prec(t_1, t_2)$ | $\forall x,y,t,t'.\ @(x, y, t)\ \&\ \prec(t,t') \Rightarrow$ @(x, y, t')  Persistence |
| | | | @(a, f, $t_2$) | | $\forall x,y,z,t.\ @(x, y, t)\ \&\ @(y, z, t) \Rightarrow$ @(x, z, t)  Transitivity |
| c. | John went to the kitchen | @(j, k, $t_3$) | @(a, j, $t_3$) @(a, k, $t_3$) | $\prec(t_2, t_3)$ | |
| d. | John dropped the apple | ¬@(a, j, $t_4$) | @(a, k, $t_4$) | $\prec(t_3, t_4)$ | contributes $-@ \otimes \mathbf{a} \otimes \mathbf{k} \otimes \mathbf{t}_4$ which cancels inference from Persistence Axiom |
| e. | Where was the apple before the kitchen? → the office | $\mathcal{O}x.\exists t,t'$. @(a, k, $t'$) & @(a, x, t) & $\prec(t, t')$ | $t' = t_2, t = t_3$ $x = f$ | | |

To answer the query, we construct its TP encoding[†]:

(13) The query

Where was the apple before the kitchen?   $\mathcal{O}x.\exists t',t.$ @(a, k, $t'$) & @(a, x, t) & $\prec(t, t', \emptyset)$

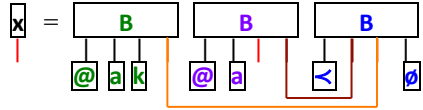

Penrose Tensor Diagram

Component-wise expression

---

[†] This tensor equation is the TP encoding of the form of the query expression given in (13); an alternative that replaces one factor of **B** with a factor $\mathbb{T}$ is the TP encoding of an alternative form of the query: $\mathcal{O}x.\exists t_i.$ @(a, k, $t_{i+1}$) & @(a, x, $t_i$). Either form of the query is slightly simplified; the additional requirement "$x \neq \underline{k}$" is needed if @(a, k, t) persists across two consecutive times. In the TP encoding, this yields an additional factor, or a simple post-processing step, that projects onto the subspace orthogonal to **k**.

## 2 SIMPLIFICATION

### 2.1 Deriving the simplification for two-place predicates

The first simplification is to omit the vector encoding the relation "is at", which we've written "@". As shown in Lee et al. (2016), for most of the bAbI problem types, this is the only relation needed (they are "uni-relational"), so it is not necessary to encode it explicitly: all items have the same initial tensor factor **@**. This simplification is shown in the third column of table (14). The matrix-algebra expression $\mathbf{xy^T}$ and the tensor expression $\mathbf{x} \otimes \mathbf{y}$ define exactly the same elements $[\mathbf{xy^T}]_{jk} = x_j y_k = [\mathbf{x} \otimes \mathbf{y}]_{jk}$.

(14) Simplification sufficient for the bAbI task: implicit is at predicate, implicit time stamps

| Symbolic | Full TPR | Simplification: 1 | 2 | 3 | 4 |
|---|---|---|---|---|---|
| $\{@(x, y, t_1),$ | $\mathbf{@} \otimes \mathbf{x} \otimes \mathbf{y} \otimes \mathbf{t}_1$ | $\mathbf{xy^T}$ | $\mathbf{x} \otimes \mathbf{y}$ | $\mathbf{x} \otimes \mathbf{y} \otimes \begin{bmatrix}1\\0\end{bmatrix} + \mathbf{y} \otimes \mathbf{z} \otimes \begin{bmatrix}0\\1\end{bmatrix}$ | $\mathbf{x} \otimes \mathbf{y} \otimes \mathbf{t}_1$ |
| $@(y, z, t_2)\}$ | $+ \mathbf{@} \otimes \mathbf{y} \otimes \mathbf{z} \otimes \mathbf{t}_2$ | $\mathbf{yz^T}$ $\cong$ | $\mathbf{y} \otimes \mathbf{z}$ $=$ | $=$ | $+ \mathbf{y} \otimes \mathbf{z} \otimes \mathbf{t}_2$ |

The second simplification is to replace the time stamps with slots in a "memory"; in table (14), these slots are shown as a vertical queue in the "Simplification: 1" column. Rather than explicitly including a final tensor factor $\mathbf{t}_i$ that encodes an explicit time stamp, we just locate the item for time $t_i$ in the $i^{\text{th}}$ position in the queue. If we think of these memory positions as locations in a vector, as shown in the "Simplification: 2" column, however, we can recognize the vector as a sum of two vectors, each the tensor product of the cell entry with a unit column vector, as spelled out in the "Simplification: 3" column. As indicated in the "Simplification: 4" column, this reduces to just the Full TPR representation (with the initial tensor factor **@** omitted) — once we identify $\mathbf{t}_1 = (1, 0), \mathbf{t}_2 = (0, 1)$. This analysis obviously trivially extends to any number of time steps.

With the two simplifications made above, the tensor implementation of inference from the Transitivity Axiom becomes:

(15) Simplified Transitive Inference operation (predicate **p** and time **t** factors not made explicit)

  a. Full form: $\mathbb{V}[\mathbf{X}, \mathbf{Y}; t]_{\pi''\alpha\beta'\gamma''} = \Sigma_{p \in \mathcal{T}} \mathbf{p}_{\pi''} (\mathbf{p}_\pi[\mathbf{X}]_{\pi\alpha\beta\gamma} \mathbf{t}_\gamma)(\mathbf{p}_{\pi'}[\mathbf{Y}]_{\pi'\beta\beta'\gamma'} \mathbf{t}_{\gamma'}) \mathbf{t}_{\gamma''}$

  b. Simplified form: $\mathbb{V}[\mathbf{X}, \mathbf{Y}]_{\alpha\beta'} = [\mathbf{X}]_{\alpha\beta}[\mathbf{Y}]_{\beta\beta'} = [XY]_{\alpha\beta'}$, i.e., simple matrix multiplication

This is exactly the form that transitive inference takes in Lee et al. (2016); e.g., in the example type-2 question discussed there, $X = \mathbf{fm^T}$ and $Y = \mathbf{mg^T}$ — for @(football, mary, t) and @(mary, garden, t) — are combined by matrix multiplication to give $XY = \mathbf{f}(\mathbf{m^T m})\mathbf{g^T} = \mathbf{fg^T}$ — for @(football, garden, t).

### 2.2 Deriving the simplification for three-place predicates

The analyses in Lee et al. (2016) of questions in categories 2, 3 and 5 involve the binding of 3 entities rather than 2. For example:

(16) The representation of "Mary travelled to the garden [from the kitchen]" is

$\mathbf{m}(\mathbf{g} \circ \mathbf{k})^\mathsf{T}$, where $\mathbf{g} \circ \mathbf{k} \equiv U[\mathbf{g}; \mathbf{k}]$; $U: \mathbb{R}^{2d} \to \mathbb{R}^d$, $U[\mathbf{g}; \mathbf{k}] \equiv R_0 \mathbf{g} + R_1 \mathbf{k}$; $R_0, R_1: \mathbb{R}^d \to \mathbb{R}^d$

Let all of the $n$ relevant entities (actors, objects, locations, etc.) $\{e_l \mid l \in 1:n\}$ be represented by unit vectors $\{\hat{\mathbf{e}}_l\}$ in $\mathbb{R}^d$, and suppose $d = 2m$, $m \in \mathbb{N}$, with $m \geq n$. (These assumptions apply to the implementation discussed in Lee et al. (2016).) Assume the generic case in which the $\{\hat{\mathbf{e}}_l\}$ are linearly independent, and let the $n$-dimensional subspace of $\mathbb{R}^d$ that they span be $E$. Let the restrictions of $R_0, R_1$ to $E$ be denoted $\underline{R}_0 \equiv R_0|_E$, $\underline{R}_1 \equiv R_1|_E$. Assume the generic case in which $\underline{R}_0, \underline{R}_1$ are non-singular, so that $\{\underline{R}\hat{\mathbf{e}}_l \equiv \hat{\mathbf{e}}_{l0}\} \subset \mathbb{R}^d$, $\{\underline{R}\hat{\mathbf{e}}_l \equiv \hat{\mathbf{e}}_{l1}\} \subset \mathbb{R}^d$ are each linearly independent sets. In order that $U$ be information-preserving, assume that these two sets are linearly independent of each other, i.e., that the union of these two sets is also a linearly independent set $\{\hat{\mathbf{e}}_{l\beta} \mid l \in 1{:}n, \beta \in 0{:}1\} \subset \mathbb{R}^d$. This is possible because $2n \leq 2m = d$.

Because the $\underline{R}_\beta$ are non-singular, there exist inverses $\underline{R}_0^{-1}$, $\underline{R}_1^{-1}$ respectively defined over span$\{\hat{\mathbf{e}}_{l0}\} \equiv S_0$, span$\{\hat{\mathbf{e}}_{l1}\} \equiv S_1$; these are linearly independent $n$-dimensional subspaces of $\mathbb{R}^d$. There is an extension $R_0^+$ of $\underline{R}_0^{-1}$ to $S \equiv \text{span}(S_0 \cup S_1) = \text{range}(U|_E)$ such that $R_0^+|_{S_0} = \underline{R}_0^{-1}$ and $R_0^+|_{S_1} = \mathbb{O}$ (namely $R_0^+\left(\Sigma_{l\beta} y_{l\beta} \hat{\mathbf{e}}_{l\beta}\right) = \Sigma_l y_{l0} \hat{\mathbf{e}}_l$; since $R_0: \Sigma_l x_l \hat{\mathbf{e}}_l \mapsto \Sigma_l x_l \hat{\mathbf{e}}_{l0}$, we get $R_0^+ R_0 = \mathbb{1}|_E$ ). Similarly there exists $R_1^+$ such that $R_1^+|_{S_1} = \underline{R}_1^{-1}$ and $R_1^+|_{S_0} = \mathbb{O}$. Thus:

(17)  Inverting ∘

$$R_0^+(\mathbf{a} \circ \mathbf{b}) = \mathbf{a},\ R_1^+(\mathbf{a} \circ \mathbf{b}) = \mathbf{b},\ \text{for all } \mathbf{a}, \mathbf{b} \in E$$

The binding in (16) can be identified as a Contracted TPR as follows. Recall that the matrix product is a kind of tensor inner product, that is, a contraction of a tensor (outer) product: $[M\mathbf{v}]_k \equiv \Sigma_j v_j M_{kj} \equiv [\mathfrak{C}_{13} \mathbf{v} \otimes \mathbf{M}]_k$ where $\mathbf{v}$ is the vector $\mathbf{v}$ considered as an order-1 tensor and $\mathbf{M}$ is the matrix $M$ considered as an order-2 tensor. In particular, $[R_0 \mathbf{a}]_k = [\mathfrak{C}_{13} \mathbf{a} \otimes \mathbf{R}_0]_k$ and $[R_1 \mathbf{a}]_k = [\mathfrak{C}_{13} \mathbf{a} \otimes \mathbf{R}_1]_k$. Thus:

(18)  The representation of "$\underline{m}_2$ travelled $\underline{\Delta}_3$; $\underline{\Delta}_3$ = to $\underline{g}_0$ from $\underline{k}_1$" as a Contracted TPR

$[\mathbf{m}\,(\mathbf{g} \circ \mathbf{k})^\mathsf{T}]_{jk} = [\mathbf{m} \otimes \boldsymbol{\Delta}]_{jk}$    "$\underline{m}_2$ travelled $\underline{\Delta}_3$": contextual-binding of fillers m, $\Delta$ of slots 2, 3

$\boldsymbol{\Delta} = \mathfrak{C}_{13}[\mathbf{g} \otimes \mathbf{R}_0 + \mathbf{k} \otimes \mathbf{R}_1]$    "$\underline{\Delta}_3$ = to $\underline{g}_0$ from $\underline{k}_1$": positional-binding of fillers g, k to roles 0, 1

$\mathbf{R}_0$ and $\mathbf{R}_1$ are the tensors representing the roles 0, 1 in "to$\underline{\ \ }_0$ from $\underline{\ \ }_1$".

Alternatively, eliminating the displacement $\Delta$, we have, for "$\underline{m}_2$ travelled to $\underline{g}_0$ from $\underline{k}_1$":

$[\mathbf{m}\,(\mathbf{g} \circ \mathbf{k})^\mathsf{T}]_{jk} = \mathfrak{C}_{24}[\mathbf{m} \otimes (\mathbf{g} \otimes \mathbf{R}_0 + \mathbf{k} \otimes \mathbf{R}_1)]_{jk}$

Unbinding the actor a by left-multiplying by $\mathbf{a}^\mathsf{T}$ gives the displacement $\Delta$ of a that is represented:

$[\mathbf{a}^\mathsf{T}\,\mathbf{m}\,(\mathbf{g} \circ \mathbf{k})^\mathsf{T}]_k = \Sigma_j [\mathbf{a}]_j\,[\mathbf{m}\,(\mathbf{g} \circ \mathbf{k})^\mathsf{T}]_{jk} = (\Sigma_j [\mathbf{a}]_j\,[\mathbf{m}]_j)\,(\mathbf{g} \circ \mathbf{k})^\mathsf{T}_k = (\mathbf{a} \cdot \mathbf{m})\,(\mathbf{g} \circ \mathbf{k})^\mathsf{T}_k$

$= (\mathbf{a} \bullet_{12} \mathfrak{C}_{24}[\mathbf{m} \otimes (\mathbf{g} \otimes \mathbf{R}_0 + \mathbf{k} \otimes \mathbf{R}_1)])_k \equiv \mathfrak{C}_{12,35}\,[\mathbf{a} \otimes \mathbf{m} \otimes (\mathbf{g} \otimes \mathbf{R}_0 + \mathbf{k} \otimes \mathbf{R}_1)]_k$

$= (\mathbf{a} \bullet_{12} \mathbf{m})\,\mathfrak{C}_{13}[\mathbf{g} \otimes \mathbf{R}_0 + \mathbf{k} \otimes \mathbf{R}_1]_k$

The factor $(\mathbf{a} \bullet_{12} \mathbf{m}) = \mathbf{a} \cdot \mathbf{m} = 1$ if $\mathbf{a} = \mathbf{m}$ (entity vectors are normalized), while $\mathbf{a} \cdot \mathbf{m}$ is considerably less than 1 if the $n$ entity vectors are generically distributed in a considerably larger space $\mathbb{R}^d$, $d \geq 2n$; indeed, we have been assuming that these $n$ vectors have been chosen to be orthogonal, in which case we have exactly $\mathbf{a} \cdot \mathbf{m} = 0$ when $\mathbf{a} \neq \mathbf{m}$. When $\mathbf{a} = \mathbf{m}$ the result is $\mathfrak{C}_{13}[\mathbf{g} \otimes \mathbf{R}_0 + \mathbf{k} \otimes \mathbf{R}_1]$, the Contracted TPR of the pair $\langle$g, k$\rangle = \Delta$: this tells us that the represented displacement of m was: to g from k. The entities g and k can be extracted from $\boldsymbol{\Delta}$ via the inner products with the duals of the role tensors, as is standard for TPRs: $\mathbf{g} = \boldsymbol{\Delta} \bullet_{13} \mathbf{R}_0^+$, $\mathbf{k} = \boldsymbol{\Delta} \bullet_{13} \mathbf{R}_1^+$; these equations are the tensor counterparts of $R_0^+(R_0\,\mathbf{g} + R_1\mathbf{k}) = \mathbf{g}$, $R_1^+(R_0\mathbf{g} + R_1\mathbf{k}) = \mathbf{k}$.

In Lee et al. (2016), questions in category 5 are treated with an operation * that functions identically to ∘, with $d \times 2d$ matrix $V$ rather than $U$. The same analysis just given for ∘/$U$ applies to show that the */$V$ method amounts to a Contracted TPR.

## 2.3 Deriving the simplification for Path Finding (bAbI category 19)

An example of a bAbI Category 19 problem is given in (19).

(19) Path-Finding: example problem

|    | Sentence $i$                              | LF: L($i$)      | Full TPR                                    | Model      |
|----|-------------------------------------------|-----------------|---------------------------------------------|------------|
| a. | The bedroom is south of the hallway.      | s(b, h)         | $\mathbf{s} \otimes \mathbf{b} \otimes \mathbf{h}$ | b = Sh     |
| b. | The bathroom is east of the office.       | e(a, o)         | $\mathbf{e} \otimes \mathbf{a} \otimes \mathbf{o}$ | a = Eo     |
| c. | The kitchen is west of the garden.        | w(k, g)         | $\mathbf{w} \otimes \mathbf{k} \otimes \mathbf{g}$ | k = Wg     |
| d. | The garden is south of the office.        | s(g, o)         | $\mathbf{s} \otimes \mathbf{g} \otimes \mathbf{o}$ | g = So     |
| e. | The office is south of the bedroom.       | s(o, b)         | $\mathbf{s} \otimes \mathbf{o} \otimes \mathbf{b}$ | o = Sb     |
| f. | How do you go from the garden to the bedroom? | ⌐P.P(g, b)  |                                             |            |
|    | → P = p[n, n]                             |                 |                                             |            |

The expression "p[s,w]" denotes "the path consisting of west then south". p accepts as an argument a list of directions, so that, in general, "p[$d_n$, …, $d_1$]" denotes the path consisting of $d_1$ (∈ {n, s, e, w}) followed by $d_2$ followed by … followed by $d_n$.

The rules of inference needed to solve such Path-Finding problems are given in (20).

(20) Axioms/rules of inference for Path-Finding problems: $\forall x,y,z,d,d_1,…,d_n,d'_1,…,d'_{n'}$ …

   a. n($x$, $y$) ⇒ s($y$, $x$); s($x$, $y$) ⇒ n($y$, $x$); e($x$, $y$) ⇒ w($y$, $x$); w($x$, $y$) ⇒ e($y$, $x$)

   b. d($x$, $y$) ⇒ p[d]($x$, $y$)

   c. p[$d_n$, …, $d_1$]($z$, $y$) & p[$d'_{n'}$, …, $d'_1$]($y$, $x$) ⇒ p[$d_n$, …, $d_1$, $d'_{n'}$, …, $d'_1$]($z$, $x$)

The rules in (20a) express the inverse semantics within the pairs n ↔ s, e ↔ w. The rule (20b) states, for example, that if $x$ is (one block) north of $y$ (on a Manhattan-like grid of locations) — n($x$, $y$) — then the path consisting of (a one-block step in the direction) north — p[n] — goes to $x$ from $y$.

Finally (20c) asserts that if p[$d_n$, …, $d_1$] — the path consisting of $d_1$ (∈ {n, s, e, w}) followed by $d_2$ followed by … followed by $d_n$ — leads to $z$ from $y$, and the path p[$d'_{n'}$, …, $d'_1$] — consisting of $d'_1$ followed by … followed by $d'_{n'}$ — leads to $y$ from $x$, then p[$d_n$, …, $d_1$, $d'_{n'}$, …, $d'_1$] leads to $z$ from $x$. This rule resembles the transitivity rule p($z$, $y$) & p($y$, $x$) ⇒ p($z$, $x$), but whereas transitivity involves only a single relation, the path-finding rule involves the productive combination of multiple relations. In this sense, Path Finding is a "multi-relational" problem, whereas all the simpler bAbI problem types, reducible to transitivity, are "uni-relational" — a main point of Lee et al. (2016).

Reasoning with the multi-relational axioms in (20) can be implemented in TPR as in the simpler case of transitive inference (5) above: (7), (11). The third column of table (19) shows the full TPR encoding of the statement of the example problem. The symbolic knowledge base $\mathcal{B}$ is a set of stated and inferred propositions $\{d_i(x_i, y_i)\} \cup \{p[P^k](w_k, z_k)\}$, with $P^k = d^k_{n_k} \cdots d^k_2 d^k_1$. The TPR of $\mathcal{B}$, $\mathbf{B}$, is the direct sum of tensors of the forms $\mathbf{d}_i \otimes \mathbf{x}_i \otimes \mathbf{y}_i$ and $\mathbf{p} \otimes \mathbf{P}^k \otimes \mathbf{w}_k \otimes \mathbf{z}_k$, where $\mathbf{P}^k \equiv \mathbf{d}^k_{n_k} \otimes \cdots \otimes \mathbf{d}^k_2 \otimes \mathbf{d}^k_1$. To reply to the query "How do you go from $u$ to $v$?" we test possible paths $P$ to see whether $P$ leads to $v$ from $u$; in the bAbI task, only paths up to length 2 need to be considered. To test whether p[$P$]($v$, $u$) ∈ $\mathcal{B}$, we take the inner product of $\mathbf{B}$ with $\mathbf{p} \otimes \mathbf{P} \otimes \mathbf{v} \otimes \mathbf{u}$; the result is 0 or 1, the truth value of p[$P$]($v$, $u$).

There is a simplification of the full TP analysis that relies on a vectorial 'model' of the axioms (20), in the sense of model theory in mathematical logic: a set of linear-algebraic objects which are inter-related in ways that satisfy the axioms. The final column of table (19) shows the corresponding representations in this simpler implementation. In the full TP analysis, the vectors encoding locations and directions are arbitrary orthonormal vectors. In the simpler model, there are systematic relations between the encodings of locations and directions. Specifically, the directions north and east are encoded by $d \times d$ matrices **N**, **E**, where locations $x$ are encoded by vectors $\mathbf{x} \in \mathbb{R}^d$. Rather than having inference relations such as (20a) implementing the inverse relationships among directions, the matrices encoding south and west are systematically related to those encoding north, east, by: $\mathbf{S} = \mathbf{N}^{-1}$, $\mathbf{W} = \mathbf{E}^{-1}$. (This requires that **N**, **E** be non-singular.) And rather than adding to **B** an arbitrary fact-tensor $\mathbf{n} \otimes \mathbf{x} \otimes \mathbf{y}$, the truth of n($x$, $y$) is encoded in the relation among the encoding vectors and matrices themselves: $\mathbf{x} = \mathbf{N}\mathbf{y}$. These conditions ensure that the vectorial encodings of positions and directions provide a model for the axioms (20a): n($x$, $y$), encoded as $\mathbf{x} = \mathbf{N}\mathbf{y}$, entails $\mathbf{y} = \mathbf{N}^{-1}\mathbf{x} = \mathbf{S}\mathbf{x}$, the encoding of s($y$, $x$).

If $L$ is the set of all possible locations for the given problems, then $\{\mathbf{N}\mathbf{y} \mid \mathbf{y} \in L\}$ and $\{\mathbf{E}\mathbf{y} \mid \mathbf{y} \in L\}$ must be independent sets of vectors, i.e., we need the following condition: range($\mathbf{N}|_L$) and range($\mathbf{E}|_L$) are linearly independent, so $2|L| \leq d$. (This is reminiscent of the conditions on $R_0$, $R_1$ in (16).)

For paths, we let the encoding of p[$d_n$, …, $d_1$]($z$, $y$) be $\mathbf{z} = \mathbf{D}_n \cdots \mathbf{D}_1 \mathbf{y}$. These encodings provide a model for the composition axiom, since the encodings of p[$d_n$, …, $d_1$]($z$, $y$) and p[$d'_{n'}$, …, $d'_1$]($y$, $x$), $\mathbf{z} = \mathbf{D}_n \cdots \mathbf{D}_1 \mathbf{y}$ and $\mathbf{y} = \mathbf{D}'_{n'} \cdots \mathbf{D}'_1 \mathbf{x}$, entail that $\mathbf{z} = \mathbf{D}_n \cdots \mathbf{D}_1 \mathbf{D}'_{n'} \cdots \mathbf{D}'_1 \mathbf{x}$, the encoding of p[$d_n$, …, $d_1$, $d'_{n'}$, …, $d'_1$]($z$, $x$). Also, the base case expressed by axiom (20b) is satisfied, since $d(x, y)$ and p[$d$]($x$, $y$) have the same encoding, $\mathbf{x} = \mathbf{D}\mathbf{y}$.

In the simplified approach implemented in Lee et al. (2016), a set of position vectors and direction matrices encoding the statements given in the problem is generated. Then, to test whether p[$P$]($v$, $u$) for a given path $P$, to determine whether $P$ answers the query "how do you go from $u$ to $v$?", the validity of the equation $\mathbf{v} = \mathbf{P}\mathbf{u}$ is determined, where $\mathbf{P} = \mathbf{D}$ or $\mathbf{D}_2 \mathbf{D}_1$ in accord with $P$ = p[$d$] or p[$d_2$, $d_1$].

## 2.4 Performance of simplification on bAbI dataset

The simplification of the full TPR reasoning analysis described above was implemented and the results reported in Lee et al. (2016) are briefly summarized in (21).

(21) Performance of the simplification on the bAbI tasks

    100% in all question categories except:

        C5:    99.8%

        C16:  99.5%

Because the present analysis performs inference by programmed vector procedures rather than learned network computations, this performance cannot be directly compared to that of previous work addressing the bAbI task (including, notably, Peng et. al (2015)'s Neural Reasoner which achieved 66.4%/97.9% and 17.3%/87.0% on tasks 17 and 19, with 1k/10k training examples, respectively; these are the most difficult tasks, on which the previous best performance was 72% and 36%). (Previous best performance on C5/C16 were 99.3%/100%, by the strongly supervised Memory Network of Weston, Chopra & Bordes (2014).)

# 3 GENERAL TREATMENT

(22) General case of query construction

a. $\sigma x_{k_x(1)}^{i_x(1)} \cdots x_{k_x(q)}^{i_x(q)} \exists e_{k_e(1)}^{i_e(1)} \cdots e_{k_e(s)}^{i_e(s)} \bigwedge_{k \in 1:n} p_k(v_k^1, \ldots, v_k^m) \bigwedge_{j \in 1:E} [e_{k_e'(j)}^{i_e'(j)} = e_{k_e''(j)}^{i_e''(j)}]$

b. $v_k^i \in \{c_k^i, x_k^i, e_k^i\}$

c. $\text{ans}_{\gamma_{k_x(1)}^{i_x(1)} \cdots \gamma_{k_x(q)}^{i_x(q)}} = \prod_{k \in 1:n} B_{\pi_k \gamma_k^1 \cdots \gamma_k^m} \prod_{i,k: v_k^i = c_k^i} [c_k^i]^{\gamma_k^i} \prod_{j \in 1:E} \delta^{\gamma_{k_e'(j)}^{i_e'(j)} \gamma_{k_e''(j)}^{i_e''(j)}}$

How the particular example query (13) follows from the general case (22) is spelled out in (23).

(23) Derivation of the query (13) from the general case (22)

a. $\sigma x. \exists t', t.\ @(a, k, t')\ \&\ @(a, x, t)\ \&\ \prec(t, t', \varnothing)$

b. $\sigma x_2^2. \exists e_1^3, e_2^3.\quad p_1(c_1^1, c_1^2, e_1^3) \qquad p_1 = @;\ c_1^1 = a,\ c_1^2 = k,\ e_1^3 = t'$
$\&\ p_2(c_2^1, x_2^2, e_2^3) \qquad p_2 = @;\ c_2^1 = a,\ v_2^2 = x,\ e_2^3 = t$
$\&\ p_3(e_3^1, e_3^2, c_3^3) \qquad p_3 = \prec;\ e_3^1 = t,\ e_3^2 = t',\ c_3^3 = \varnothing$
$\&\ [e_1^3 = e_3^2]\ \&\ [e_2^3 = e_3^1]$

c. $\text{ans}_{\gamma_2^2} = B_{\pi_1 \gamma_1^1 \gamma_1^2 \gamma_1^3} B_{\pi_2 \gamma_2^1 \gamma_2^2 \gamma_2^3} B_{\pi_3 \gamma_3^1 \gamma_3^2 \gamma_3^3} [c_1^1]_{\gamma_1^1} [c_1^2]_{\gamma_1^2} [c_2^1]_{\gamma_2^1} [c_3^3]_{\gamma_3^3} \delta^{\gamma_1^3 \gamma_3^2} \delta^{\gamma_2^3 \gamma_3^1}$

d. $x_{\beta_2} = B_{\pi_1 \alpha_1 \beta_1 \gamma_1} B_{\pi_2 \alpha_2 \beta_2 \gamma_2} B_{\pi_3 \alpha_3 \beta_3 \gamma_3} a_{\alpha_1} k_{\beta_1} a_{\alpha_2} \varnothing_{\gamma_3} \delta^{\gamma_1 \beta_3} \delta^{\gamma_2 \alpha_3}$

Analogous methods allow general TP instantiation of rules of inference, from which the particular forms in (7)–(8) can similarly be derived.